# A mutli-thread tabu search algorithm

A.M. Connor

*Engineering Design Centre, Department of Engineering,
University of Cambridge, Cambridge, UK*

**Abstract** This paper describes a novel refinement to a Tabu search algorithm that has been implemented in an attempt to improve the robustness of the search when applied to particularly complex problems. In this approach, two Tabu searches are carried out in parallel. Each search thread is characterised by it's own short term memory which forces that point out of local optima. However, the two search threads share an intermediate term memory so allowing a degree of information to be passed between them. Results are presented for both unconstrained and constrained numerical functions as well as a problem in the field of hydraulic circuit optimization. Simulation of hydraulic circuit performance is achieved by linking the optimization algorithm to the commercial simulation package Bathfp.

**Introduction**

Tabu search (Glover, 1989, 1990) is a metaheuristic procedure which is intended to guide optimization methods to avoid local optima during complex or multi-modal numerical optimization problems. Tabu search has previously been applied to a variety of classical and practical problems which include the quadratic assignment problem (Skorin-Kapov, 1989), electronic circuit design (Bland and Dawson, 1991), the balancing of hydraulic turbines (Sinclair, 1993), design of structural components (Hu, 1992; Bland, 1994) and radar polyphase code design (Cangalovic et al, 1996). A complete coverage of the method and a survey of applications can be found in the literature (Glover and Laguna, 1997).

The Tabu search method utilises a number of flexible memory cycles, each with different associated time scales, to allow search information to be exploited more thoroughly than by rigid memory or memory-less systems.

Previous work (Connor and Tilley, 1997, 1998a, 1998b) in the field of hydraulic circuit optimization has shown that the aggressive nature of Tabu search has many advantages over other novel optimization techniques. The main advantage has been the reduction in the number of circuit simulations required to find a feasible solution. However, this work suffers from one of the drawbacks of earlier work (Donne, 1993) in this area which utilised a parallel Genetic Algorithm in that both methods lack robustness and subsequent attempts at the same problem often lead to different feasible solutions.

This is caused by a number of factors which are related to the complexity of the problem which cause a degree of lack of definition in the objective function. The optimization of hydraulic circuits involves complex multi-criteria objective functions and often large numbers of penalty function terms to make sure that the solution obtained matches the specified design requirements. Work to improve upon objective function definition is currently being undertaken, but the following method has been developed to achieve greater robustness and consistency.

A brief description of the standard Tabu search algorithm will be given before later sections describe the multi-thread approach.

**Tabu Search Algorithms**

The power of the Tabu search algorithm is derived from the use of flexible memory cycles of differing time spans. These memory cycles are used to control, intensify and diversify the search in order to find a suitable solution.

*Short Term Memory*

The most simple implementation of a Tabu search is based around the use of a hill climbing algorithm. Once the method has located a locally optimal solution, the short term memory is used to force the search out from this optima in a different direction.

The short term memory constitutes a form of aggressive search that seeks to always make the best allowable move from any given position. To achieve an appropriate balance between effort expended to identify a best move (where "best" can be defined by multiple and varying criteria), candidate list strategies are characteristically used to reduce the number of moves to be considered at each iteration.

Short term memory is implemented in the form of *tabu restrictions* which prevent the search from cycling around a given optimal or sub-optimal position, and to impart vigour to the search. Essentially, these restrictions apply to a limited set of previously visited positions, and the restrictions are continuously updated as the search progresses. In one form of implementation, attributes associated with each new trial solution are placed in the memory list when the solution is generated and corresponding attributes of the oldest solution in the list are removed so that the size of the list remains constant. Other implementations use dynamic lists whose size vary in different phases of the search.

The short term memory cycle has been implemented in the described work as a list of tabu restrictions. The tabu list contains the parameter values of the last *n* accepted solutions. During the search, before the objective function of a proposed point is evaluated, its parameter set is compared to those contained in the tabu list. If a match is found, disclosing that a solution embodying these same parameters has been recently visited, then the objective function is not evaluated and the search classes the move as Tabu.

In many Tabu search implementations the tabu restrictions are allowed to be rescinded if appropriate aspiration criteria are satisfied. For example, a criterion called "aspiration by objective" allows an otherwise tabu solution to be visited if its objective function value is better than that of any solution previously obtained (or any previously obtained that shares selected features in common with the current solution). Aspiration criteria have not been included in this work, though they could be incorporated in the algorithm.

*Search Intensification*

Intensification of the search is based around the concept of the intermediate memory cycle. In this case, the intermediate memory is very similar to the short term memory in that it is a list of solution attributes. However, the solutions contained in the intermediate memory list are the previous *m* best solutions. As a new best solution is found, this is placed into the list and the worst solution removed. This is termed intermediate memory as the time scale of replacement is much longer than for short term memory.

The intermediate memory list contains information about the best solutions located so far and so allows similarities between good solutions to be extracted and used to generate new solutions with good characteristics.

*Search Diversification*

In many Tabu search applications, search diversification is achieved by the use of long term memory cycles or by refreshing the current base point, either randomly or strategically, when certain conditions apply. However, in previous work relating to hydraulic circuit optimization certain parallels have been drawn to the notion of schema processing within Genetic Algorithms to propose a novel diversification strategy that is intended to force the search towards new areas of the search domain which may also have good objective function values. The search is refreshed by randomly selecting a value for each search parameter from the solutions located in the intermediate memory list. This selection is not limited to selecting values from the same parameter of the solutions, so in effect any value of any parameter can be selected for the newly proposed solution.

This diversification strategy allows values that are good for one parameter to be tried for a different parameter and is particularly effective when all of the design parameters have similar ranges or are normalised in some way. This concept is reinforced by the Tabu search principle of creating new attributes that represent usefully exploitable problem features. In this case, the new attributes may be conceived of as "parameter classes" where all members of a class may be treated in similar way.

*Hill Climbing Algorithm*

The underlying search algorithm used in this work is a two stage hill climbing search (Hooke and Jeeves, 1961). The first stage carries out an initial exploration around a given base point by increasing and decreasing each design variable by a given amount known as the step size. Once this exploration has been carried out and a new point found, the search is extended along the same vector by a factor *k*. This is known as a pattern move. If the pattern move locates a solution with a better objective value than the exploration point, then this point is used as a new base point and the search is repeated. Otherwise, the search is repeated using the exploration point as a new base point.

In the standard Hooke and Jeeves implementation, this process is repeated until no further improvement is found. When this occurs, the step size is reduced and the search continued. This is repeated until the step size falls below a given value. In the Tabu search, this strategy is somewhat modified. Firstly, the selection of the new base point is adjusted so that the search always selects the best available move. When no improvement can be found, the best available move is defined as that where the increase in objective function value is the smallest. Because of this, a new control algorithm has been developed which reduces the step size as the search progresses and eventually terminates the search. This algorithm is shown in flow chart form in Figure 1.

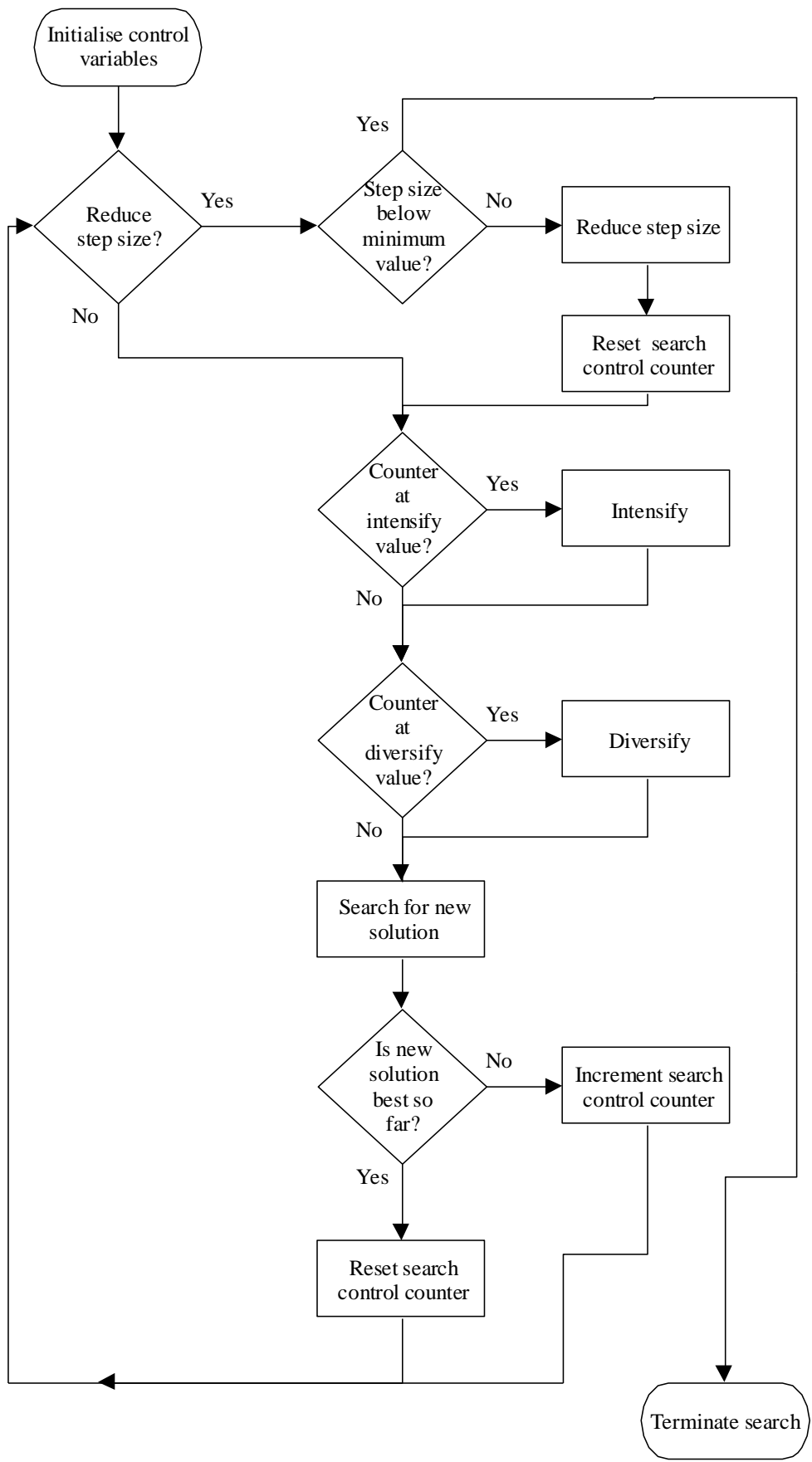

Figure 1. Search Algorithm

*Mutli-Thread Tabu Search*

The multi-thread implementation is an attempt to improve the robustness of the Tabu search concept by introducing a degree of parallelism into the search. The current implementation utilises two search threads which are initialised at different start points.

Each search thread has it's own short term memory (tabu list) which is used to force each point out of local optima. However, the two search threads share a common intermediate memory so that information can be exchanged between the two points during intensification and diversification of the search. The approach of sharing access to elite solutions in a parallel Tabu search implementation has also been utilised in a number of other studies (Toulouse, Crainic and Gendreau, 1996; Crainic, Toulouse and Gendrau, 1997).

The implementation in this work uses the same controlling algorithm as the single thread approach, but differs slightly in that only one thread undergoes either intensification or diversification at any one stage in the search. This considerably reduces the chance of the two threads ever representing the same trial solution. If this occurs, then the two threads will then follow the same trajectory through the search space. Future implementations could eliminate this possibility by comparing the individual short term memory lists at various stage throughout the search. Should the tabu restrictions be identical, then the two threads could be forced apart by some means.

**Numerical Test Functions**

Two numerical functions have been used to assess the performance of the multi-thread search, with respect to the performance of a single thread Tabu search algorithm. These are the Schwefel function and the Bump function.

*Schwefel Function*

The Scwhefel function is a very complex ten parameter function with a unique minimum and very large numbers of local minima in the search region. Previous work (Donne, 1993) has suggested that methods capable of optimising the Schwefel function are likely to be suitable for optimising hydraulic circuits. The objective function is calculated using Equation 1.

$$obfn = \sum_{i=1}^{10} x_i \sin\left(\sqrt{abs(x_i)}\right) \tag{1}$$

The search variables are limited to the region enclosed by the bounds $-500 \leq x_i \leq 500$ and minimum step size for each variable is 0.0001. The function does not encompass any constraints other than the upper and lower boundaries. The unique minimum of the function is located at $x_i = 420.9687$ and has a value of -4189.83.

Table I shows the results from five runs of both the standard and multi-thread Tabu search methods.

**Table I. Results for the Schwefel Function**

| RUN | SINGLE POINT OBFN VALUE | SINGLE POINT NO. EVALS | MUTLI-THREAD OBFN VALUE | MULTI-THREAD NO. EVALS |
|---|---|---|---|---|
| 1 | -4189.83 | 12032 | -4189.83 | 28806 |
| 2 | -4189.83 | 11732 | -4189.83 | 25551 |
| 3 | -4189.83 | 9194 | -4189.83 | 23337 |
| 4 | -4189.83 | 10783 | -4189.83 | 18009 |
| 5 | -4189.83 | 10310 | -4189.83 | 23198 |

As can be seen, in all cases the methods have located the optimum value. However, the multi-thread method has required a greater number of evaluations. This is because the same termination criteria have been applied, but the method uses two search points. The multi-thread method has no advantages over the standard method for this function.

*Bump Function*

The Bump function (Keane, 1994) is a considerably more complex function to optimise than the Schwefel function. It produces a very complex function surface for any number of variables. This study attempts two variations on this function, where the number of variables are set to 20 and 50. The objective function value is calculated using equation 2.

$$obfn = \frac{\sum_{i=1}^{n} \cos^4 x_i - 2\left(\cos^2 x_1 \times \cos^2 x_2 \times \cdots \times \cos^2 x_n\right)}{\sum_{i=1}^{n} x_i^2 i} \quad (2)$$

The function is further complicated in that it is a constrained by two inequality constraints and that optimum solutions generally lie very near the constraint boundary. The inequality constraints are given in equation 3 and equation 4.

$$\prod_{i=1}^{n} x_i > 0.75 \quad (3)$$

$$\sum_{i=1}^{n} x_i < \frac{15n}{2} \quad (4)$$

The search variables are limited to the region enclosed by the bounds $0 \leq x_i \leq 10$. Again, the minimum step size for each parameter has been set to 0.0001. The maximum of the function is not known, but appears to be around 0.82 for the n=50 problem (Keane, 1994).

In the Bump function it is recommended that the search be initialised from the starting point where $x_i=5$. In this work, results are presented for the single thread implementation for this case and also when the start point is selected at random. For the multi-thread method, one of the search threads is started from the recommended point whilst the other is randomly generated.

Table II shows the results from five runs of the single thread Tabu search for when n=20. Results are shown for both the seeded and the random start.

**Table II. Results for Single Thread Method on Bump Function (n=20)**

| RUN | SINGLE POINT OBFN VALUE (SEEDED) | SINGLE POINT NO. EVALS (SEEDED) | SINGLE POINT OBFN VALUE (RANDOM) | SINGLE POINT NO. EVALS (RANDOM) |
|---|---|---|---|---|
| 1 | 0.65495 | 7493 | 0.579423 | 10689 |
| 2 | 0.65495 | 7485 | 0.549577 | 14842 |
| 3 | 0.65495 | 7491 | 0.653184 | 8985 |
| 4 | 0.65495 | 7456 | 0.584064 | 6547 |
| 5 | 0.65495 | 7481 | 0.722475 | 13921 |

Table III shows the results obtained using the multi-thread method where one search thread is started from the recommended values and the other is started from a randomly generated point.

**Table III. Results for Mutli-Thread Method on Bump Function (n=20)**

| RUN | MUTLI-THREAD OBFN VALUE | MUTLI-THREAD NO. EVALS |
|---|---|---|
| 1 | 0.781727 | 13214 |
| 2 | 0.729619 | 14183 |
| 3 | 0.606138 | 11862 |
| 4 | 0.759619 | 12494 |
| 5 | 0.781727 | 13941 |

Table IV shows the results from five runs of the single thread Tabu search methods for when n=50. Results are shown for both the seeded and the random start.

**Table IV. Results for Single Thread Method on Bump Function (n=50)**

| RUN | SINGLE POINT OBFN VALUE (SEEDED) | SINGLE POINT NO. EVALS (SEEDED) | SINGLE POINT OBFN VALUE (RANDOM) | SINGLE POINT NO. EVALS (RANDOM) |
|---|---|---|---|---|
| 1 | 0.664774 | 24428 | 0.617665 | 30314 |
| 2 | 0.664774 | 24454 | 0.525611 | 38879 |
| 3 | 0.662558 | 22909 | 0.385927 | 30283 |
| 4 | 0.664774 | 24435 | 0.620488 | 30334 |
| 5 | 0.664774 | 24454 | 0.588570 | 33651 |

Table V shows the results obtained using the multi-thread method where one search thread is started from the recommended values and the other is started from a randomly generated point.

**Table V. Results for Mutli-Thread Method on Bump Function (n=50)**

| RUN | MUTLI-THREAD OBFN VALUE | MUTLI-THREAD NO. EVALS |
|---|---|---|
| 1 | 0.808646 | 49146 |
| 2 | 0.778217 | 55111 |
| 3 | 0.758960 | 48287 |
| 4 | 0.758845 | 49074 |
| 5 | 0.758845 | 49168 |

For both of the Bump functions the single thread implementation, when initialised from a fixed point, is very consistent which suggest that the current degree of probability in the diversification strategy does not effect the final converged value to any great extent. However, when started from a random start point, the final values obtained do not exhibit such consistency.

For both of the Bump functions the multi-thread Tabu search is locating solutions with better objective function values than the single thread approach. This indicates that the search is inherently more diverse due to the parallelisation that has been introduced. This increases the possibility of locating the true global optimum as the search is exploring more than one area of the search space at a given time.

**Fluid Power Circuit Optimization**

A number of numerical optimisation approaches have previously been used to automatically size and select components in fluid power circuits by linking the algorithm to the fluid power simulation environment Bathfp (Tomlinson & Tilley, 1993). The algorithms used include Genetic Algorithms (Donne, 1993) and Tabu search (Connor and Tilley, 1998). Bathfp utilises a complex type insensitive numerical integrator (Richards et al, 1990) based upon the Lsoda algorithm (Petzold, 1983) which changes the time step and integration method depending on the stiffness of the differential equations that describe the system being simulated.

Whilst the Lsoda algorithm provides an accurate method of simulating the performance of fluid power circuits, including those which have discontinuous properties, the actual simulation of complex circuits can take considerable lengths of time. It is not uncommon to take several hours to simulate a few minutes of real time. Due to this, previous attempts using a parallel Genetic Algorithm have been considered unsuccessful as typically they involved around 50-60,000 evaluations to find a suitable solution to even the most simple of circuit problems. This number of evaluations would typically take several days running on a Sun SPARC 20 workstation. This has sparked an interest in the use of Tabu search which has been claimed in the literature (Sinclair, 1993; Borup and Parkinson, 1992) to be considerably more efficient than other optimisation methods.

The two Tabu search algorithms described in this paper have been tested on a number of problems in the field of hydraulic circuit optimization. Previous work (Donne, 1993; Donne *et al*, 1994; Connor and Tilley, 1998a) has shown that the following circuit is particularly difficult to optimise. In this circuit the design objectives are to have two motors operating at two different speeds, but driven from the same pump. This is achieved by the use of pressure compensated flow valves (PCFV) and the required circuit is shown in Figure 2.

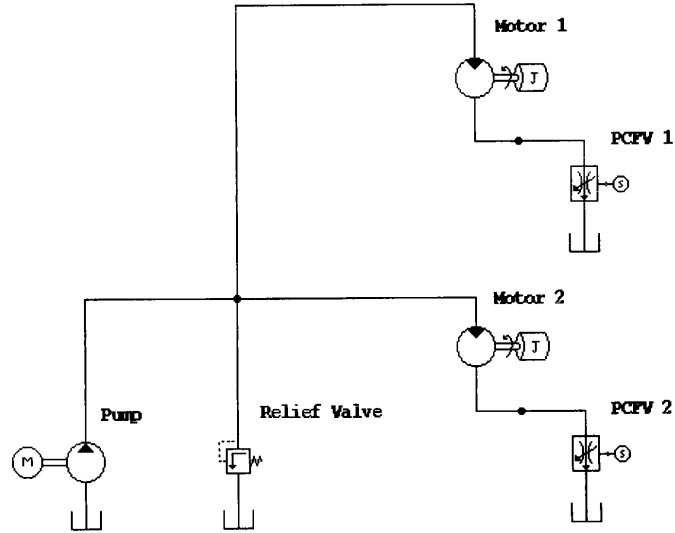

**Fig 2. Test Hydraulic Circuit**

In this example, the design variables selected for optimization are the displacements of the pump and both motors, along with the nominal flow rates of the two control valves. The pump and motor displacements are constrained between 1cc/rev and 1000 cc/rev. The nominal flow rate settings of the PCFVs are constrained between 10 and 100 L/min. The desired speeds for the two motors are 120 rpm and 60 rpm. The objective function is given in equation 5.

$$obfn = \{e(\omega_1)^2 + e(\omega_2)^2\} \times \left\{1 + \frac{Q_{RV}}{Q_P}\right\} \quad (5)$$

In this equation, the values used to calculate the objective function value are the value of the state variables at the end of the simulated run period. The error in each of the motor speeds, $e(\omega_1)$ and $e(\omega_2)$, is squared to penalise high speed errors. The sum of the errors is then further penalised by multiplying by a normalised value that expresses the percentage of the flow that is produced by the pump, $Q_P$, that is returning to the tank through the relief valve, $Q_{RV}$. Table VI shows the results for the single thread Tabu search on this problem and Table VII shows the results for the multi-thread method.

**Table VI. Results for Single Thread Method**

| RUN | PUMP SIZE (CC/REV) | MOTOR 1 SIZE (CC/REV) | MOTOR 2 SIZE (CC/REV) | PCFV 1 FLOWRATE (L/MIN) | PCFV 2 FLOWRATE (L/MIN) | NO. EVALS | OBFN VALUE |
|---|---|---|---|---|---|---|---|
| 1 | 35 | 129 | 609 | 31 | 31 | 1848 | 0.0442421 |
| 2 | 63 | 369 | 830 | 100 | 48 | 1423 | 0.126669 |
| 3 | 41 | 139 | 738 | 16 | 38 | 1405 | 0.28996 |
| 4 | 70 | 74 | 195 | 91 | 11 | 1442 | 0.0917881 |
| 5 | 65 | 363 | 880 | 48 | 51 | 1091 | 1.93982 |
| 6 | 9 | 90 | 39 | 9 | 2 | 1337 | 0.989931 |
| 7 | 33 | 304 | 211 | 66 | 12 | 1592 | 0.0304883 |
| 8 | 50 | 153 | 936 | 100 | 51 | 1470 | 0.170034 |
| 9 | 65 | 579 | 460 | 70 | 27 | 1494 | 0.065249 |
| 10 | 61 | 405 | 708 | 100 | 41 | 1583 | 0.0540956 |
| | | | | | | Average | 0.380228 |

**Table VII. Results for Mutli-Thread Method**

| RUN | PUMP SIZE (CC/REV) | MOTOR 1 SIZE (CC/REV) | MOTOR 2 SIZE (CC/REV) | PCFV 1 FLOWRATE (L/MIN) | PCFV 2 FLOWRATE (L/MIN) | NO. EVALS | OBFN VALUE |
|---|---|---|---|---|---|---|---|
| 1 | 44 | 221 | 650 | 76 | 36 | 2872 | 0.0847743 |
| 2 | 74 | 511 | 820 | 64 | 48 | 2570 | 0.0627989 |
| 3 | 69 | 418 | 831 | 49 | 47 | 2410 | 0.0530253 |
| 4 | 48 | 297 | 601 | 100 | 34 | 2968 | 0.357169 |
| 5 | 50 | 449 | 334 | 65 | 20 | 2948 | 0.205888 |
| 6 | 20 | 203 | 137 | 21 | 4 | 2106 | 0.134311 |
| 7 | 44 | 400 | 294 | 58 | 17 | 2304 | 0.0536162 |
| 8 | 72 | 599 | 596 | 82 | 35 | 2500 | 0.0900985 |
| 9 | 61 | 544 | 424 | 100 | 25 | 2406 | 0.898487 |
| 10 | 51 | 170 | 927 | 59 | 51 | 3928 | 0.00566847 |
| | | | | | | Average | 0.19458367 |

As the absolute minimum objective function value is not known, these results can be analysed by considering the actual parameters obtained as well as the objective function value. The first point to notice is that for both methods, there is a significant spread of parameter values. This indicates the interactive nature of hydraulic circuits in which different parameter size combinations can produce the same performance. These different parameter sets often have similar objective function values. However, it can be seen that the multi-thread Tabu search is locating solutions which have, on average, lower objective function values than those located by the single point method.

Table VIII shows the values of the desired state variables at the end of the simulated time period for the best solutions located by the two methods.

**Table VIII. Motor Speeds and Relief Valve Flows**

|  | SINGLE | MULTI |
|---|---|---|
| $\omega_1$ (RPM) | 119.97 | 119.998 |
| $\omega_2$ (RPM) | 59.9992 | 59.9965 |
| $Q_{RV}$ (L/MIN) | 0 | 0 |
| OBFN VALUE | 0.03049 | 0.00567 |

The speed error in both cases is insignificant which indicates that further work is required to highlight design objectives with in hydraulic circuit optimization which would allow for better objective function definition. However, as the complexity of the objective function is increased it becomes increasingly difficult to produce an acceptable compounded objective function and it is possible to advocate a move to a true multi-objective optimisation approach.

**Conclusions**

This paper has described how the a Tabu search algorithm can be adapted to deal with multiple search threads which exchange information concerning the solutions space through a common intermediate memory cycle.

This multi-thread Tabu search appears to be more robust than the single thread implementation on well defined numerical problems. This is not so apparent on real problems in hydraulic circuit optimization due to the poor objective function definition and interactive nature of hydraulic circuits. This interaction allows solutions with different parameter sets to have similar objective function values. Despite this, the multi-thread search is locating solutions which have, on average, lower objective function values than the solutions located by a single search thread.

More work is obviously required in encapsulated knowledge concerning the design objectives into the objective function, but the results presented in this paper indicate that the multi-thread Tabu search shows considerable promise as an optimization strategy. The method is both reasonably consistent and is locating solutions with a relatively low number of objective function evaluations.